\documentclass{article}

  \usepackage[preprint,nonatbib]{neurips_2025}

\usepackage[utf8]{inputenc} %
\usepackage[T1]{fontenc}    %
\usepackage{booktabs}       %
\usepackage{amsfonts}       %
\usepackage{nicefrac}       %
\usepackage{microtype}      %
\usepackage{xcolor}         %
\usepackage{antonio}
\usepackage{wrapfig}

\title{Deep Sensorimotor Control by \\ Imitating Predictive Models of Human Motion}

\author{Himanshu Gaurav Singh$^1$, Pieter Abbeel$^1$, Jitendra Malik$^1$, Antonio Loquercio$^2$ \\
$^1$UC Berkeley \quad $^2$University of Pennsylvania }

\begin{document}

\maketitle

\begin{abstract}
As the embodiment gap between a robot and a human narrows, new opportunities arise to leverage datasets of humans interacting with their surroundings for robot learning.
We propose a novel technique for training sensorimotor policies with reinforcement learning by imitating predictive models of human motions.
Our key insight is that the motion of keypoints on human-inspired robot end-effectors closely mirrors the motion of corresponding human body keypoints.
This enables us to use a model trained to predict future motion on human data \emph{zero-shot} on robot data.
We train sensorimotor policies to track the predictions of such a model, conditioned on a history of past robot states, while optimizing a relatively sparse task reward.
This approach entirely bypasses gradient-based kinematic retargeting and adversarial losses, which limit existing methods from fully leveraging the scale and diversity of modern human-scene interaction datasets.
Empirically, we find that our approach can work across robots and tasks, outperforming existing baselines by a large margin.
In addition, we find that tracking a human motion model can substitute for carefully designed dense rewards and curricula in manipulation tasks. Code, data and qualitative results available at \url{https://jirl-upenn.github.io/track_reward/}
\end{abstract}

\section{Introduction}
\label{intro}

Training robot policies using datasets of humans interacting with their surroundings is a promising approach to scaling robot learning.
Indeed, there is no shortage of such datasets, thanks to recent advances in 3D vision~\cite{pavlakos2024reconstructing,goel2023humans,wu20244d,kerbl20233d}, industrial augmented and virtual reality devices, \eg, Meta Oculus or Apple Vision Pro, and specifically designed tooling~\cite{guzov2024hmd,ma2024nymeria}.
However, leveraging these datasets to train effective sensorimotor robot policies remains an open challenge.

A common strategy to address this challenge is \emph{kinematic retargeting}, which maps human motions to a robot’s embodiment. This process is typically performed \emph{independently for each sample} via gradient-based optimization, subject to the robot’s kinematic and dynamic constraints. The resulting sensorimotor trajectories can then be used to (pre-)train robot policies through imitation learning~\cite{qin2022dexmv,singh2024hand,radosavovic2024humanoid,ye2023learning,shaw2023videodex}. However, kinematic retargeting methods often \emph{assume access to a simulated replica of the environment} to evaluate feasibility, e.g., collision checking, which in turn requires accurate 3D scene reconstruction. These requirements are particularly challenging for manipulation tasks, which involve contact-rich interactions and dynamic, non-static scenes.
While recent techniques address some of these issues~\cite{lakshmipathy2023contact,lakshmipathy2024kinematic}, they still need to make scene and/or task-specific approximations to the full robot dynamics. Because of these factors, retargeting methods require significant effort when mapping large human-interaction datasets.

Another approach to training robot policies from datasets of humans interacting with a scene is demonstration-guided reinforcement learning (RL).
One line of work~\cite{peng2020learning, peng2021amp, escontrela2022adversarial, qin2022dexmv} uses adversarial rewards to match state distributions between the human and the robot. However, adversarial objectives are unstable to train and are therefore generally employed within small-scale and well-curated datasets.
Another line of work combines online policy gradient with imitation losses on offline expert trajectories~\cite{rajeswaran2017learning, ball2023efficient}.
However, to be used for gradient computation, expert trajectories need to contain robot action annotations, which are absent in datasets of human activities and can only be recovered via retargeting.

\begin{figure*}
    \centering
    \includegraphics[width=0.95\linewidth]{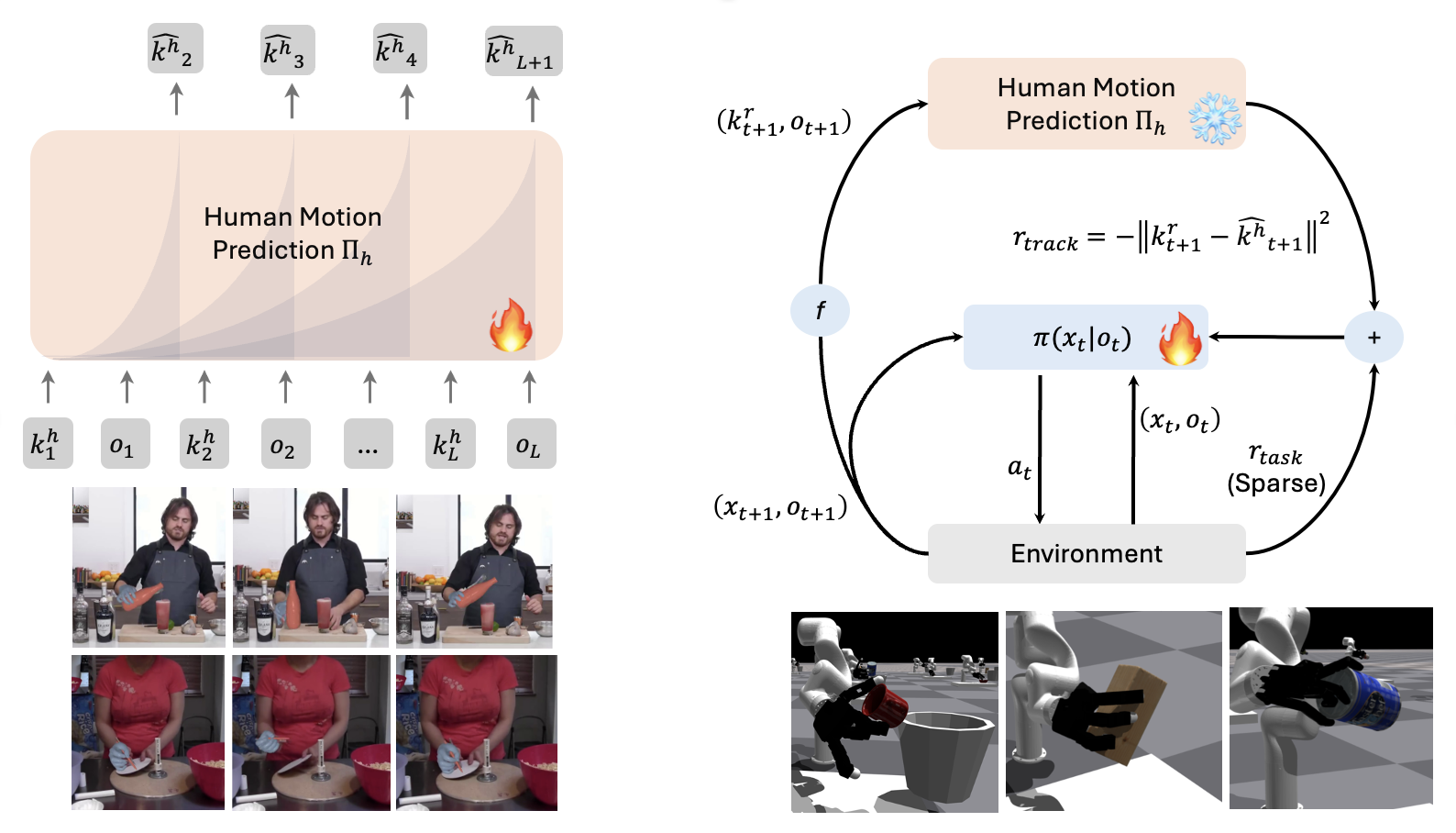}
    \caption{\textbf{Method.} We use a dataset of humans interacting with their scene to train a motion prediction model $\Pi_h$. Such model, instantiated as a causal transformer, takes as input a history of previous 3D keypoints \(k^h_{t:t-L}\), \ie, the location of the human's fingertips, and observations $o_{t:t-L}$, \ie, the objects' pointcloud, to predict the future human keypoint location $\hat{k}^h_{t+1}$. For anthropomorphic robots, thanks to the abstraction of keypoints, $\Pi_h$ can be used on robot data despite being trained on human data. Therefore, $\Pi_h$ can predict likely human motions while training a policy $\pi_{\theta}$ on a downstream task. A reward to track such motions, $r_{track}$, provides an additional training signal to the otherwise sparse task reward $r_{task}$.}
    \label{fig:method}
\end{figure*}

In this paper, we introduce a simple yet scalable approach for learning from datasets of humans interacting with their environment. We revisit the core observation of kinematic-retargeting based approaches: modern robotic embodiments are often designed to mimic human form factors~\cite{ciocarlie2009hand}. As a result, the motion of keypoints on the end-effectors of human-inspired robots closely resembles the motion of keypoints on the human body~\cite{ciocarlie2007dimensionality}, despite fundamental differences in their action spaces.  We use this observation for a novel insight: a predictive model trained on human keypoints can be applied \emph{zero-shot} to robot data. First, we train a predictive model, \(\Pi_h\), to estimate the future locations of keypoints on the human body based on a history of previous scene observations and keypoint locations. Then, we use RL in simulation to maximize a task reward while tracking the predicted future human keypoints. These predictions are obtained by feeding a history of the robot’s observations and keypoint locations into \(\Pi_h\).  Figure~\ref{fig:method} provides an overview of our approach.

In contrast to the conventional paradigm, our approach offers several key advantages:
(i) It removes the need to replicate the environment in which humans interact within a simulator. Indeed, generalization of the predictive model to the scene in simulation removes the need for accurate \textit{real-to-sim}. (ii) It decouples human data from the robot policy learning loop, meaning the potentially large human dataset is not required during policy training; (iii) It enables the automatic selection of the most suitable skill within the human dataset’s simplex to solve the task at hand, eliminating the need for manually defined selection procedures~\cite{papagiannis2024r+, mandikal2022dexvip}.

Similar to previous approaches, our method requires a mapping between human and robot keypoints (see Figure~\ref{fig:mapping}). However, we find that this mapping is straightforward to design and remains effective across different tasks and embodiments.  

Our experiments demonstrate that incorporating a reward for tracking predictive models of human motion enables robots to learn dexterous manipulation tasks from relatively sparse task rewards, eliminating the need for carefully engineered reward functions.
Moreover, we show that the motion predictor $\Pi_h$ can be applied \emph{zero-shot} across different tasks and embodiments, allowing our approach to outperform existing baselines.

Overall, our approach serves as a stepping stone toward fully leveraging the richness of existing datasets of human activities for sensorimotor policy learning.

\section{Related Work}
\label{sec:rel_work}

As the embodiment gap between a robot and a human closes, the amount of information that can be learned via observation increases~\cite{ciocarlie2007dimensionality}, as well as the sources of data robots can learn from~\cite{romero2011human}. In the next paragraphs, we will provide an overview of different ways to learn from human data and map human motion to robot motion using kinematic constraints and/or adversarial losses.

\subsection{Cross-Embodiment Transfer}

A natural approach to leveraging human data for robot training is to find an abstraction of human motion that eliminates embodiment-specific factors. For robots with an anthropomorphic hand, one such abstraction is the 3D location of the hand's wrist and fingertips~\cite{garcia2020physics, antotsiou2018task, mandikal2022dexvip, qin2022dexmv, singh2024hand, ye2023learning}. Selecting fewer keypoints, \eg only the wrist location, enables transfer between different kinds of end-effectors. However, this often requires additional demonstrations for the transfer between human and robot motion to be effective~\cite{kareer2024egomimic,ren2025motion,wang2023mimicplay}.

Instead of focusing on end-effectors, an alternative approach to map motion between embodiments is to focus on the effects of actions—such as the objects being manipulated~\cite{chen2024object}. This perspective allows for transfer between vastly different embodiments. Indeed, the only requirement is that the outcomes of the actions remain consistent.

A common and data-efficient approach to achieving this is through affordances, which indicate where an object can be interacted with~\cite{bahl2023affordances, mandikal2020dexterous, wu2023learning}. However, while affordance-based methods are data-efficient, they need to handle pre- and post-contact trajectories separately with specifically designed modules. One way to address this limitation is by modeling the object's 3D~\cite{kokic2020learning} or 2D motion~\cite{yuan2024general, xu2024flow}.

Incorporating further contextual cues can add constraints to improve the transfer between embodiment. This can be done by modeling the two-dimensional motion of the entire scene~\cite{liu2018imitation, bahl2022human, bharadhwaj2024track2act}. Importantly, mapping human motion and object motion are not mutually exclusive; they can be combined to maintain data efficiency while improving effectiveness for non-anthropomorphic robots~\cite{papagiannis2024r+, heppert2024ditto, Wen2023AnypointTM}.

The mapping between keypoints on the human's and robot's end effectors does not necessarily need to be manually defined, but could be learned from data. A common approach to do so is by transforming the entire human demonstration to a robot demonstration via a generative model and map the 2D motion of the rendered robot to actions via reinforcement learning~\cite{smith2019avid,xiong2021learning}. Instead of rendering the video with a different embodiment, another possibility is generating either the full video~\cite{bharadhwaj2024gen2act} or a segmented version of it~\cite{bharadhwaj2024towards} from only the initial observation. Such generated video can then be translated into robot actions by collecting a set of paired demonstrations.

Similarly to our work, these approaches leverage the idea that the data of humans interacting with their surroundings can be translated into sensorimotor robot trajectories. However, they make the additional assumption that the scene between the human and the robot matches. In addition, the mapping is usually learned from data and does not account for the robot kinematic and dynamic constraints. In the next section, we will cover a family of methods that explicitly account for such constraints during motion mapping.

\subsection{Kinematic Retargeting}

The mapping of human motion to robot motion and actions has traditionally been cast as a constraint optimization problem. Such optimization, however, is challenging when the human interacts with the scene, \eg, picking up an object, as contact dynamics is non-differentiable. Finding effective solutions to this problem has long been studied~\cite{liu2009dextrous,ye2012synthesis}, but it is still an active area of research~\cite{lakshmipathy2024kinematic,lakshmipathy2023contact}.

Retargeting noisy data acquired from videos adds another layer of complexity, deriving from the noise in humans and objects' pose estimation. One approach to do so is disregarding the robot's and/or objects' dynamics and recovering successful trajectories by sampling~\cite{antotsiou2018task,ye2023learning,chen2022dextransfer} or reinforcement learning~\cite{qin2022dexmv,peng2018deepmimic,peng2018sfv}. Other works include the dynamics directly in the optimization~\cite{singh2024hand}, which, however, is only feasible when a fast dynamics simulator is available. 

Such optimization becomes more challenging when mapping not only end-effectors, \eg, hands, but the whole body~\cite{villegas2021contact,kim2016retargeting}, or the interactions between different bodies~\cite{jin2018aura}. One approach to finding solutions is imposing a reduced structure to the problem, which makes it amenable to neural network optimization~\cite{aberman2020skeleton}. However, by simplifying part of the problem, \eg, ignoring the whole-body kinematics and the scene's dynamics, enables retargeting to be used for real-world whole-body control. Such simplifications can be accounted for either with reinforcement learning~\cite{ji2024exbody2} or via a human teleoperator~\cite{cheng2024tv,fu2024humanplus}.

A limitation of constrained-based optimization approaches is that motion is mapped from the human to the robot \emph{per sample}. To be effective in cases where there is interaction with a scene, a replica of the environment the human is interacting with is required to constrain the optimization. Our approach proposes a novel way to address this limitation by decoupling human data from robot policy learning.

\subsection{Adversarial Losses for Motion Matching}

Another way to match the distribution of human and robot states and observations \emph{in expectation} is via adversarial losses~\cite{torabi2018generative}. 
This approach has been very successful for character animation~\cite{barsoum2018hp,peng2021amp,li2022ganimator}, but has also been applied to robotics for manipulation~\cite{garcia2020physics} and locomotion~\cite{peng2020learning,escontrela2022adversarial}.
However, a downside of adversarial losses is that they are challenging to optimize on large datasets, which makes this line of work difficult to scale. 
Combining adversarial losses with a predictive model of human motion via control hierarchies can address this limitation~\cite{tevet2024closd}. However, while this was successful for character animation, it still presents challenges when applied in robotics, as downstream controllers might not be available in the first place for the task at hand.

\section{Preliminaries}

We adopt the standard approach of defining a sensorimotor control task as a discrete-time, finite-horizon, discounted Markov decision process (MDP), represented by the tuple \( M = (\mathbb{S}, \mathbb{A}, \mathbb{P}, r, \rho_0, \gamma, T) \). Here, \( \mathbb{S} \) is the state space, \( \mathbb{A} \) is the action space, \( \mathbb{P}: \mathbb{S} \times \mathbb{A} \times \mathbb{S} \to \mathbb{R} \) is the transition probability distribution, \( r: \mathbb{S} \times \mathbb{A} \to \mathbb{R} \) is the reward function, \( \rho_0: \mathbb{S} \to \mathbb{R} \) is the initial state distribution, \( \gamma \) is the discount factor, and \( T \) is the time horizon.

The objective is to optimize a stochastic policy \( \pi_\theta: \mathbb{S} \times \mathbb{A} \to \mathbb{A} \), parameterized by \( \theta \), by maximizing its discounted expected return 
$R(\pi_\theta) = \mathbb{E}_\tau \left[ \sum_{t=0}^T \gamma^t r(s_t, a_t) \right]$,
where \( \tau = (s_0, a_0, \ldots) \) denotes the trajectory of states, actions, and goals encountered during an episode. Specifically, \( s_0 \sim \rho_0 \) is the initial state sampled from the state distribution, \( a_t \sim \pi_\theta(a_t | s_t) \) is the action sampled from the policy, and \( s_{t+1} \sim \mathbb{P}(s_{t+1} | s_t, a_t) \) is the next state sampled from the transition distribution.

\section{Method}

\begin{figure*}
    \centering
    \includegraphics[width=\linewidth]{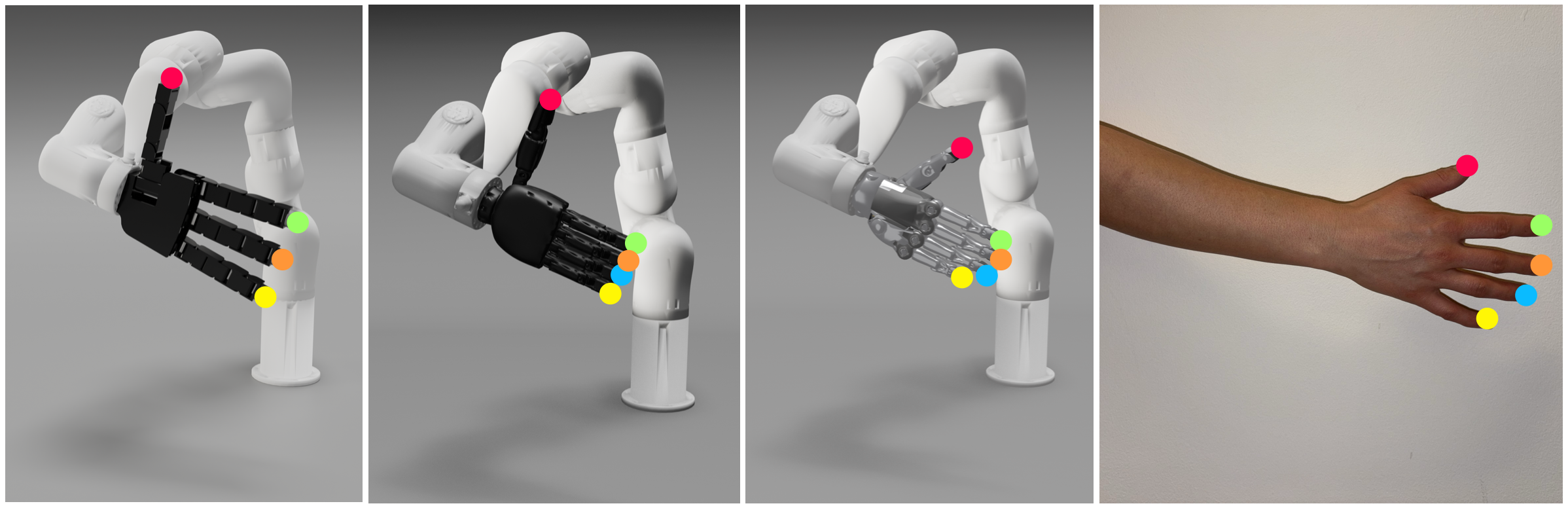}
    \caption{\textbf{Mapping between human and robot hands} Anthropomorphic hands allow an intuitive mapping of robot links to human hand keypoints. It is defined once and remains consistent across tasks. Here, we show the mapping of the human hand to three different morphologies: an Allegro hand \cite{allegro} (left), an Xhand \cite{xhand} (center), and an SVH hand \cite{svh}(right).}
    \label{fig:mapping}
\end{figure*}

Tasks with a high-dimensional state and action space often require the reward to be carefully crafted to achieve the desired behavior.
Motivated by the challenges of designing rewards for complex tasks, we propose a simple approach to bias the agent towards behaviors likely to be performed by humans.
Specifically, we shape a sparse task reward \(r_{task}\) by incorporating an additive term \(r_{track}\), which incentivizes the robot to follow the predicted motion of a human while doing the same task.

To derive \(r_{track}\), we assume the existence of an abstraction level at which the difference between human and robot motion is minimal. For anthropomorphic robots and manipulation tasks, such abstraction is easy to define (Fig.~\ref{fig:mapping}).
Following prior work on kinematic retargeting~\cite{ciocarlie2007dimensionality,qin2022dexmv,shaw2023videodex,ye2023learning,singh2024hand}, we define this abstraction as the 3D positions of fingertips.

The reward \(r_{track}\) measures the distance between the robot and 3D human keypoints, \ie, 
\begin{equation}
    r_{track} = -\| k^h_{t+1} - f(s_{t+1}) \|^2=-\| k^h_{t+1} - k^r_{t+1} \|^2,
    \label{eq:rh}
\end{equation}
where $k^h_{t+1} \in \mathbb{B}$ are the keypoints 3D location on the human hand, and $f: \mathbb{S} \mapsto \mathbb{B}$ is the function to compute the robot's keypoints location $k^r_{t+1}$ from its state, \ie, forward kinematics. 

Conceptually, Eq.~\ref{eq:rh} is straightforward and has been explored in prior works that learn to translate human behavior into that of robot \emph{per-demonstration}~\cite{peng2018deepmimic, peng2020learning, mandikal2022dexvip}. However, its application requires human data in an environment and task which \emph{closely matches} that of the robot. This limiting assumption can hinder its application, especially in contact-rich tasks where a suitably accurate re-construction of the human environment is hard.

To address this limitation, we propose to compute an estimate of $k^h_{t+1}$ by training a model of human motion $\Pi_{h}$ to predict the most likely future keypoints location given a history of previous keypoints and observations.
The key idea of our approach is that, thanks to the abstraction of keypoints, such a model can be trained on human data but evaluated on robot data.
Therefore, we can modify Eq.~\ref{eq:rh} to let the policy $\pi_{\theta}$ track not the observed human keypoints $k^h_{t+1}$ but the \emph{predicted} keypoints $\hat{k}^h_{t+1}$, generated by passing the robot's state history as an input to $\Pi_h$. Formally, 
\begin{equation}
    r_{track} = -\| \hat{k}^h_{t+1} - k^r_{t+1} \|^2,
    \label{eq:rh_hat}
\end{equation}
where $\hat{k}^h_{t+1}=\Pi(f(s_{t}), \ldots, f(s_{t-L}))$, with $L$ being the model's time horizon.

\subsection{Human Motion Predictor}

Before introducing the details of our approach, we clarify the meaning of state for our setting. We divide the robot's state into two components, \ie, $s_t=[x_t, o_t]$, where $x_t \in \mathbb{R}^{d}$ is the robot's proprioceptive state and $o_t \in \mathbb{R}^{100*n_o \times 3}$ is a point cloud of the objects in the scene, $n_o$ being the number of objects. For experiments, $d$ ranges from 20-30 and $n_o$ from 1 to 3. The human keypoints $k^h_t \in \mathbb{R}^{3\times d}$ are the 3D positions of the $d$  hand and keypoints in the world frame, with $d$ varying according to the robot end-effector morphology.

To train the human motion predictor $\Pi_{h}$, we assume access to a dataset $\mathcal{D} := \{g^{(i)}, (k^{h(i)}_1, o^{(i)}_1, k^{h(i)}_2, k^{h(i)}_2, \dots)\}_{i=1}^n$, which consists of trajectories of human keypoints $k^{h}_t$ and objects' pointcloud $o_t$, as well as a goal label $g$. This dataset contains instances of humans interacting with their surroundings to accomplish the task $g$.

We instantiate $\Pi_{h}$ as a vanilla transformer with causal attention. The transformer has six layers and eight heads. We tokenize keypoints $k^h_t$ with a linear transformation to a 512-dimensional hidden size. The pointcloud $o_t$ is tokenized to the same hidden size by a PointNet encoder~\cite{qi2017pointnet}. We use a context length of $L=16$, which we find to be sufficient for our setting.

We train $\Pi_{h}$ on $\mathcal{D}$ with the mean squared error loss
\[
\mathcal{L} = \mathbb{E}_{\tau \sim \mathcal{D}, \,g, o, k^h \sim \tau} \big[\|\Pi_{h}(g, o_{t:t-K}, k^h_{t:t-K}) -  k^h_{t+1})\|^2\big].
\]
We also experimented with incorporating an additional loss on the observation tokens but found that it provided only marginal benefits. To maintain simplicity, we excluded it from our final model.

To efficiently train $\Pi_h$, we employ teacher forcing. However, we find that its naïve application leads to poor performance on closed-loop robot trajectories, since autoregressive inference drifts out of the training distribution over time. Empirically, we observe that this issue can be mitigated by injecting zero-mean Gaussian noise into the input keypoints $k^h_t$. Importantly, the noise is added only to the input keypoints, and not the targets, ensuring that the model’s predictions remain precise.

\subsection{Policy Optimization}

We train specialized robot policies, \(\pi_{\theta}\), using model-free reinforcement learning on a set of downstream tasks.  
For simplicity and efficiency, we implement \(\pi_{\theta}\) as a multi-layer perceptron (MLP) with four layers and a hidden dimension of 128, which is randomly initialized before training. 
The observation space consists of a point cloud representation of the scene, which is preprocessed by a PointNet encoder, as well as the robot's proprioceptive state.

We use PPO~\cite{schulman2017proximal} as our reinforcement learning algorithm. This choice is motivated by the observation that PPO is the standard choice for a wide variety of robotics applications~\cite{kaufmann2023champion,lee2020learning,ji2024exbody2,qi2023general,handa2023dextreme}. 
We employ an asymmetric actor-critic architecture for optimization. Specifically, we give as extra information to the value function the distance between the fingers and the target's center of mass.

\section{Experimental Setup}
\label{sec:exp_setup}

Our experiments are conducted in the IsaacGym simulator~\cite{makoviychuk2021isaac} with a PhysX backend, enabling fast, parallel rigid-body physics simulation. Our experiments span three manipulation tasks and three different robot platforms.

We adapt tasks from the IsaacGymEnvs~\cite{petrenko2023dexpbt} benchmark suite to evaluate our approach. Specifically, we test on the following tasks: 
\begin{itemize}
    \item \textbf{Grasp and Lift} The robot must grasp an object and hold it at a specified height.
    \item \textbf{Grasp and Lift - Clutter} The goal robot must grasp the specified object amongts multiple distractor objects.
    \item \textbf{Lift and Throw} The agent must first lift the object and then accurately drop it into an adjacent receptacle.
    \item \textbf{Cabinet} This is a dexterous manipulation task in which a robot has to interact with an articulated object. Specifically, the agent has to open the drawer of a cabinet by grasping the handle and pulling it horizontally.
\end{itemize}
More details about task setup are available in the supplementary.
We test our approach on three robotic platforms: (i) a 4-fingered Allegro Hand \cite{allegro} with 16 degrees of freedom (DoF), (ii) a 5-fingered Xhand \cite{shadow} with 12 DoF, and (iii) a 5-fingered SVH Hand \cite{svh} with 20 DoF.
All hands are mounted on a 7-DoF Xarm7 robotic arm.

Note that we use the same \(\Pi_h\) \emph{for all robots and tasks}.
Note that we condition \(\Pi_h\) to a categorical label which identifies the object of interest.
The mapping between the human and robot hand's keyponts is also quite straightforward and fixed across tasks and embodiments.
It corresponds to a fingertip-to-fingertip matching. A visualization of the robot and mapping is available in Figure~\ref{fig:mapping}.

We use the DexYCB dataset~\cite{Chao2021DexYCBAB} to train our predictive human motion model. It comprises 1,000 trajectories of human-object interactions in cluttered environments containing up to five objects, with a total of 20 distinct object categories. 

\section{Results}

We design an experimental procedure to answer the following questions: i) Can our approach enable policy learning with relative sparse task rewards? How does it compare to carefully shaped reward functions? ii) Can the same human prediction model $\Pi_h$ be used across robot embodiments and across tasks? iii) How does our method compare to other demonstration-guided RL approaches using human data (possibly after kinematic retargeting)?

\subsection{Can tracking $\Pi_h$ make up for carefully engineered rewards?}
\label{sec:exp:tasks}

\begin{wrapfigure}{r}{0.52\textwidth}  
  \centering
  \includegraphics[width=0.5\textwidth]{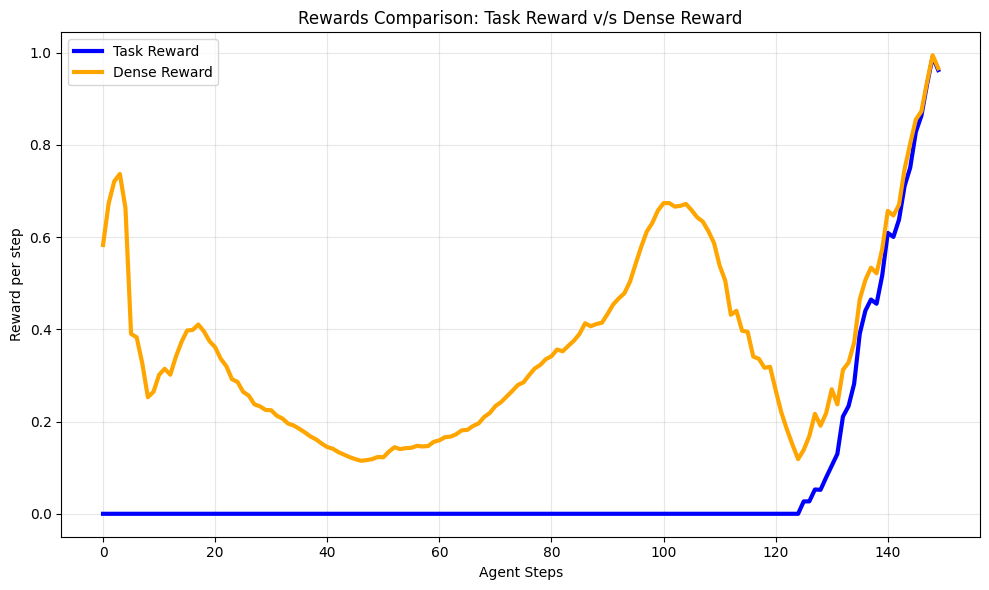}
  \caption{Comparison of sparse and dense step-wise rewards for the \textit{Grasp and Lift task} over expert policy rollouts. The sparse reward activates only after the object is grasped and moves toward the target ($\approx$step 130), while the dense reward provides continuous shaping throughout.}
  \vspace{-2ex}
  \label{fig:rewards}
\end{wrapfigure}

Reward design for robot control involves careful tuning \textit{per-task}. 
Especially for object manipulation, carefully shaped dense rewards often contain a state machine to switch between different behaviors~\cite{petrenko2023dexpbt}. Such task-specific, hand-designed procedure presents a primary bottleneck to scaling RL methods. We now 
investigate whether the tracking reward $r_{track}$ can substitute for manual reward shaping. 
To do so, we compare with the following approaches:

\begin{enumerate}
    \item \textbf{PPO-DenseReward}: We utilize the manually engineered dense reward functions provided in IsaacGymEnvs~\cite{makoviychuk2021isaac} for \textit{Grasp and Lift}, \textit{Lift and Throw}, and \textit{Open Cabinet} tasks. This reward function is a weighted sum of different components. The different reward terms are designed to: guide the hand toward the object, provide a bonus for lifting it a few centimeters, encourage movement toward the goal location after lifting, and provide a success bonus. Note that this reward function is still used in state-of-the-art reinforcement learning works~\cite{singlasapg}. We use this dense reward as a privileged baseline.
    \item \textbf{PPO-TaskReward}: This baseline uses PPO training solely with a sparse goal reward ($r_{rask}$). For \textit{Grasp and Lift} and \textit{Lift and Throw} tasks, this reward is: 
\[
\begin{aligned}
r_{rask} = (obj_z > 0.1)*\,\mathrm{ReLU}\bigl(d_t(\mathrm{goal}, \mathrm{obj}) - d_{t-1}(\mathrm{goal}, \mathrm{obj})\bigr)
\end{aligned}
\]
 where $d_t(goal,obj)$ is the distance of the object from the goal at time $t$, and $obj_z$ is the height of the object from the table. A comparison of this reward to the dense reward is shown in Fig.~\ref{fig:rewards}. For the \textit{Open Cabinet} task, the reward is
\[
\begin{aligned}
r_{rask} = drawer_x*(1 + c*ReLU(drawer_x >= drawer_{limit}))
\end{aligned}
\]
where $drawer_x$ is the distance to which the drawer body is pulled out and $drawer_{limit}$ is the distance to be pulled for the drawer to be opened.
\item \textbf{Ours}: Our method combines trajectory tracking and sparse goal rewards, \ie, using $r_{track} + \lambda*r_{task}$ as the reward function, where $\lambda$ is a hyperparameter fixed across tasks.
\end{enumerate}

\begin{figure*}[t]
    \centering
    \includegraphics[width=0.95\textwidth]{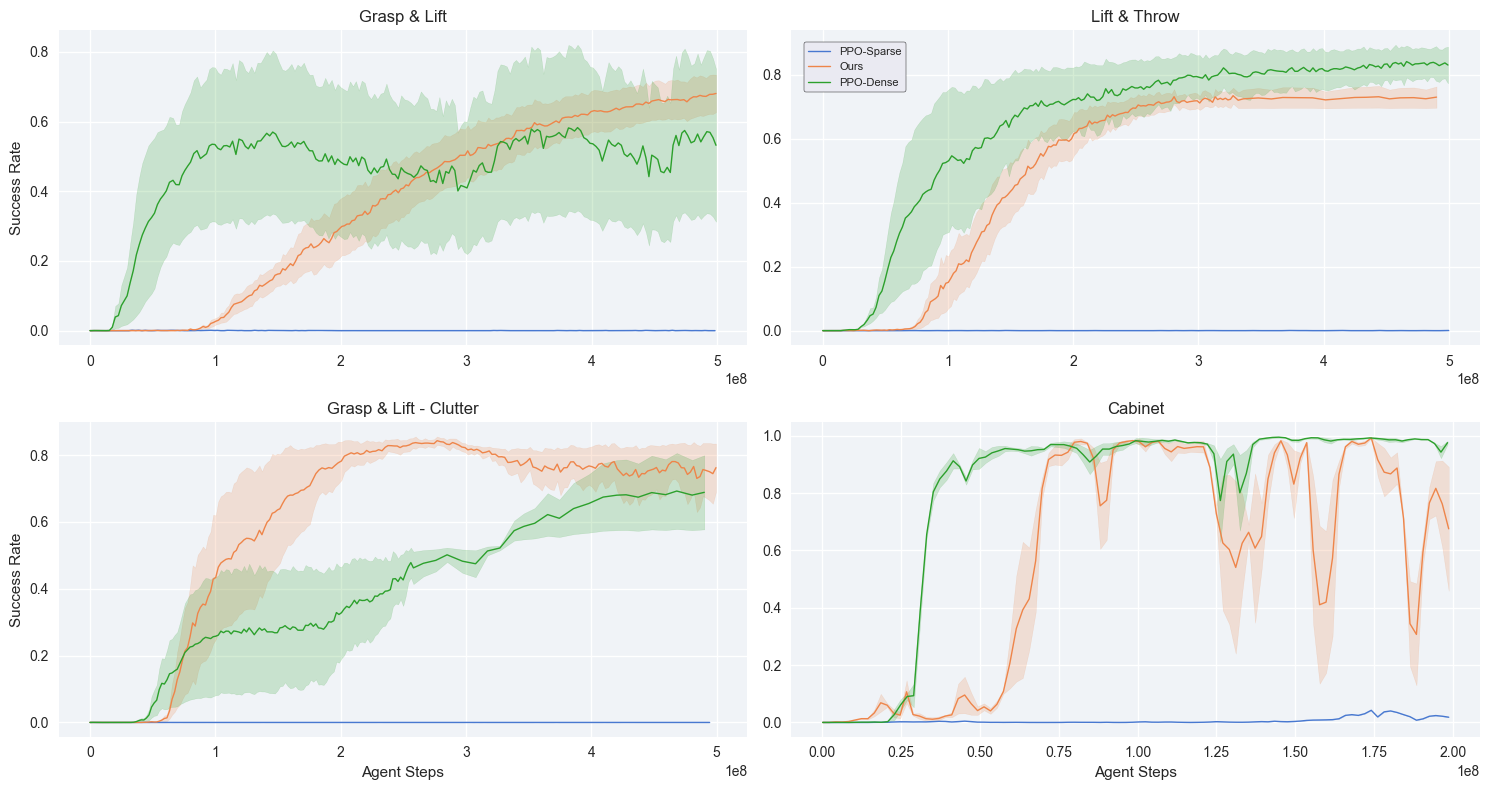}
    \caption{\footnotesize{\textbf{Comparison across tasks} Tracking the predictions of $\Pi_{h}$ enables tractable reinforcement learning with a sparse task reward. Our approach is comparable to the privileged baseline: \textit{PPO with dense rewards} across all tasks, whereas PPO with the sparse task reward only fails to learn. Runs are averaged across 3 seeds.}}
    \label{fig:exp1}
    \vspace{-1ex}
\end{figure*}

All baselines have an additional reward term penalizing the energy consumed by moving, which we approximate as the absolute sum of all joint velocities. 
We control for all additional factors, including the model architecture, agent steps, and domain randomization, but tune each baseline independently.

Figure \ref{fig:exp1} illustrates the result of this experiment. We find that our approach achieves comparable performance with respect to a highly engineered reward function.
Conversely, PPO fails to find good solutions when trained exclusively on a sparse task reward.

A more detailed breakdown of task-specific performance reveals several notable observations.
First, in the \emph{Grasp and Lift} task, our approach and PPO-Dense achieve similar performance at convergence, though PPO-Dense reaches higher success rates slightly faster.
Second, despite the absence of demonstrations for the \emph{Lift and Throw} task in the human dataset, our approach still attains a high success rate, though it performs slightly worse than PPO-Dense.
In the \emph{Grasp and Lift-Clutter} task, both methods converge to comparable performance. However, our approach solves the task more quickly, as PPO-Dense requires additional steps to distinguish the target object from distractors—knowledge that our approach automatically inherits from $\Pi_h$. For the \textit{Open Cabinet} task too, our approach performs comparably to PPO-Dense. 

\begin{figure*}
    \centering
    \includegraphics[width=\linewidth]{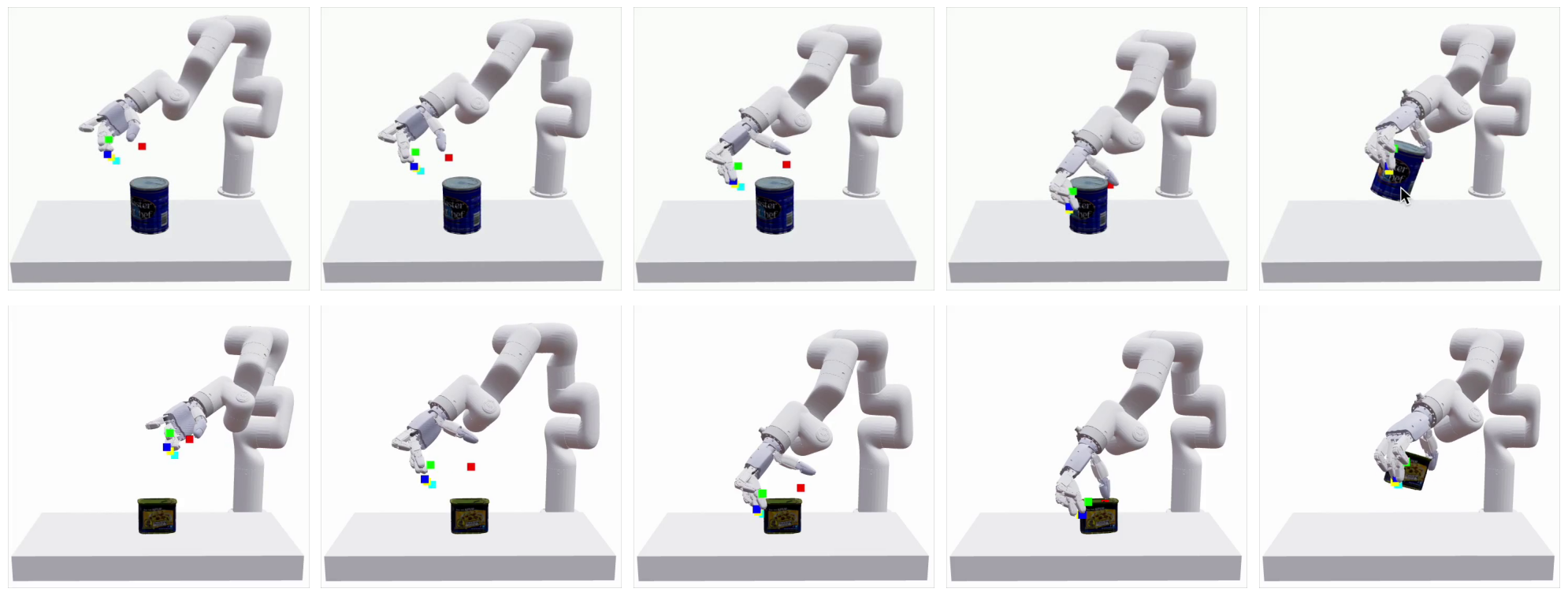}
    \caption{\textbf{Qualitative Results.} A policy trained with our approach successfully picking up two objects. The colored points show the next predictions of $\Pi_{h}$. Such predictions are temporally smooth and guide the policy towards fast grasping and lifting. Better seen at our webpage \aadd{\url{https://jirl-upenn.github.io/track_reward/}}.
}
    \label{fig:qual}
\end{figure*}

Fig.~\ref{fig:qual} shows qualitative results of our approach. It also illustrates closed-loop predictions from $\Pi_h$ based on the trajectory of robot keypoints. These predictions are key to training a policy which is smooth and effective. We refer the reader to the supplementary videos for clearer visualization of the policy behavior.

\subsection{Performance across robot embodiments}

\begin{figure*}[t]
    \includegraphics[width=0.95\textwidth]{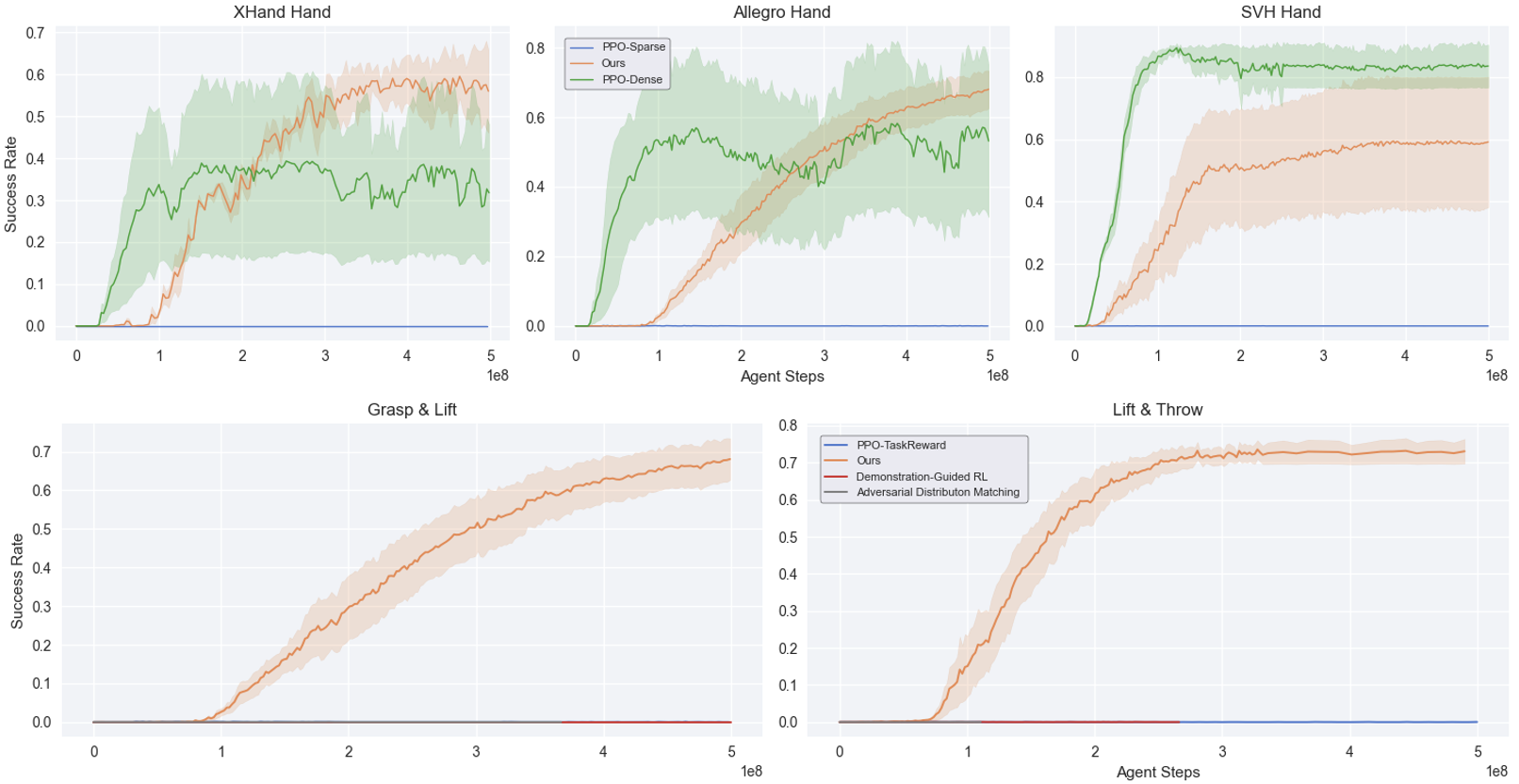}
    \caption{\footnotesize{(top) We evaluate our approach on a diverse set of multi-fingered hand embodiments. Our approach performs comparably to the privileged baseline:\textit{PPO with dense rewards} across three such robots. (bottom) Our approach outperforms existing methods that use human data to guide RL exploration, when both are trained with sparse task reward.}}
    \label{fig:multiembodiments}
\end{figure*}

In this section, we examine whether a human motion predictor can be used to provide tracking rewards across different robot embodiments. To explore this, we fix the task to \emph{Grasp and Lift} while varying the embodiment.

The results of this experiment are shown in Figure~\ref{fig:multiembodiments}. Note that we make slight changes to the PPO hyperparameters per embodiment. Specifically, we decrease the maximum learning rate in the adaptive scheduler of PPO as the DOFs in the embodiment increase. This is required because the KL-divergence between policy updates is proportional to the dimension of the action space. We attempt to do the same tuning process for the PPO-TaskReward, but we find it to be unable to learn. In contrast, our approach consistently achieves high success rates, with performance remaining stable across different embodiments.

\subsection{Comparison to baselines}

There exists a large body of work that studied the problem of using human data for training robot policies (Sec.~\ref{sec:rel_work}). From this vast literature, we select a few baselines that are compatible with our problem formulation and setup.
Specifically, we focus on approaches that have the following characteristics:
(i) They do not divide the task into subroutines, \eg, affordance detection, pre-grasp, and post-grasp motion planning; (ii) they do not require retrieving a specific example from the human dataset and mapping that to robot motion; (iii) they do not require robot demonstrations on the downstream task; (iv) they produce a policy that can be run in isolation from any other training component.
Such a filtering process is designed to make the comparison to our approach as fair as possible.

As a result of this filtering process, we compare our approach against the following baselines:
\begin{itemize}
    \item \textbf{Adversarial Distribution Matching.} This baseline trains a policy jointly with a discriminator that distinguishes between trajectory rollouts generated by the policy and those from the offline dataset. An auxiliary reward is added to the sparse task reward $r_{\text{task}}$, encouraging the policy to \textit{fool} the discriminator. In our setting, since the dataset consists of human keypoint trajectories, the discriminator is trained to differentiate between mapped robot keypoint trajectories and human keypoint trajectories. This baseline aligns with prior work on adversarial state distribution matching~\cite{torabi2018generative, peng2021amp, qin2022dexmv, escontrela2022adversarial, tevet2024closd}.
    
    \item \textbf{Retargeting, followed by demonstration-guided RL} This baseline first applies kinematic retargeting to map the human dataset into a set of sensorimotor robot trajectories. A policy is then trained by jointly optimizing $r_{task}$ on on-policy experiences and a supervised learning objective on the retargeted dataset, achieved by assigning high advantage values to these trajectories. This approach represents prior work on combining on-policy learning with off-policy trajectories derived from retargeting of human demonstrations~\cite{peng2018deepmimic, peng2018sfv, qin2022dexmv, chen2022dextransfer, ye2023learning, singh2024hand}.
\end{itemize}

As in all our previous experiments, we tune each baseline individually. Figure~\ref{fig:multiembodiments} illustrates our results. We find that the baselines fail to leverage the dataset to sufficiently bias the RL exploration required for learning from sparse rewards. 
While we found this result to be surprising, we would like to stress that the baselines are generally trained on downstream tasks with dense rewards and not previously tested in our setting. Indeed, when we experimented with providing them with dense rewards, we found them to achieve high success rates (\aadd{See Supplementary \ref{supple:densebaselines}}).

\section{Conclusions}

In this work, we studied how datasets of humans interacting with their environment can be leveraged to train robot policies. While the availability of such datasets continues to grow due to technological advancements, it remains unclear how to effectively extract and utilize this information for policy learning.
We demonstrated that a simple modification to a well-established procedure can yield significant benefits. Our approach showed promising results in dexterous manipulation across various robots and tasks. Notably, it enabled successful reinforcement learning in tasks that previously required extensive reward shaping to achieve comparable performance.

\paragraph{Limitations.}Our approach shares one key assumption of kinematic retargeting methods, \ie, that a set of keypoints on the human body and on a robot follow similar trajectories, despite significant differences in their action spaces. This limits our work to anthropomorphic robots.
While prior work has shown that clever mapping schemes can enable transfer even to robots with significantly different morphologies~\cite{ren2025motion,kareer2024egomimic,xiong2021learning}, they still require task-specific robot demonstrations to bridge the embodiment gap. 
Another limitation of this work is the lack of real-world experiments. While important, our core contribution is orthogonal to sim-to-real transfer, which typically relies on techniques such as domain randomization and sensor calibration. One hypothesis is that mimicking human motions could potentially easen the engineering load of sim-to-real transfer, since the policy inherently learns behaviours feasible in the real world. We believe this to be an interesting line of inquiry for future work.

\section{Broader Impact}
This paper presents work that aims to advance the fields of Machine Learning and Robotics. There are many potential
societal consequences of our work, none of which we feel must be specifically highlighted here.

\section{Acknowledgements}
This work was supported by the DARPA Machine Common Sense program, the DARPA Transfer from Imprecise and Abstract Models to Autonomous Technologies (TIAMAT) program, the ONR MURI award N00014-21-1-2801 and the InnoHK Clusters of the Hong Kong SAR Government via the Hong Kong Centre for
Logistics Robotics. This work was also funded by ONR MURI N00014-22-1-2773. We thank Arthur Allshire, Ankur Handa,  Jathushan Rajasegaran and Tara Sadjadpour for helpful discussions. We thank Lorenzo Bianchi and Noemi Aepli for their help with illustrations. Pieter Abbeel holds concurrent appointments as a Professor at UC Berkeley and as an Amazon Scholar. This paper describes work performed at UC Berkeley and is not associated with Amazon.

\bibliographystyle{IEEEtran}
\bibliography{antonio}
\newpage

\section{Supplementary Material}

\subsection{Analysing the behavior learnt by our policies}

Visually inspecting the behaviours learnt by the agent provides more insights into the approach. Policy rollouts in simulation for our approach can be found at \url{https://jirl-upenn.github.io/track_reward/}. The video rollouts bring out the following insights on our approach:
\begin{enumerate}
    \item \textbf{How does our prediction model guide policy training?} Predictions of our keypoint model direct the robot hand towards reasonable grasp poses of the object.
    \item \textbf{What is the impact of the hand form factor on learning?} While all robots successfully complete the task, the motion of the robot is more humanlike for hands whose form factor more closely resembles that of humans, such as the X-Hand.  
    \item  \textbf{How does our approach perform on multiple tasks?} With the right sparse task reward, our prediction model can guide the robot to do multiple, although similar, tasks.
\end{enumerate}

\subsection{{Performance of baselines with dense rewards}}
\label{supple:densebaselines}

\begin{wraptable}{r}{0.45\textwidth} %
\vspace{-0.5cm} %
\centering
\renewcommand{\arraystretch}{1.2}
\begin{tabular}{l|c}
\hline
\rule{0pt}{2.6ex} 
\textbf{Approach} & \textbf{Success Rate} \\
\hline
AMP with dense reward & $0.193 \pm 0.051$ \\
\hline
DAPG with dense reward & $0.381 \pm 0.223$ \\
\hline
\end{tabular}
\caption{Baseline performance with dense rewards.}
\label{tab:ampdense}
\vspace{-1.2em} 
\end{wraptable}

We find that our baselines, AMP \cite{peng2021amp} and Demo-guided RL \cite{rajeswaran2017learning}, that leverage offline datasets alongside policy gradients fail to learn just with sparse reward. We therefore run these methods with our manually designed dense reward function. Table~\ref{tab:ampdense} summarises our results. We find that with dense rewards, our baselines indeed lead to non-trivial success rates. We hypothesize that this reliance on dense rewards comes from the underlying policy gradient based optimization, which is known to perform worse in sparse reward environments.

\subsection{Description of our tasks in simulation}

\begin{itemize}
    \item \textit{Grasp and Lift}: The robot must grasp an object and hold it at a specified height for at least 5 seconds. The scene contains a single object, which is randomly placed on a \(1m \times 1m\) table. The episode is considered a failure if the object falls off the table, if the robot fails to maintain it at the goal height for the required duration, or if it takes more than 30 seconds to complete the task. Note that this is the simplest task in our benchmark.
    \item \textit{Grasp and Lift - Clutter}: The goal of this task is similar to the previous one: grasping an object and holding it at a specific height for at least 2.5 seconds. However, instead of a single object, three objects are randomly placed on the table. The agent must correctly identify and pick the target object, specified via a categorical label, while avoiding the other objects, which serve as distractors. The success criteria remain the same as in the previous task. 
    \item \textit{Lift and Throw}: In this task, the agent must first lift the object and then accurately drop it into an adjacent receptacle. The episode is considered a failure if the agent fails to place the object in the bin or exceeds the 30-second time limit. Since our human dataset does not contain instances of grasp-and-drop actions, this task is designed to evaluate whether a policy can still be trained for tasks \(\Pi_h\) was not explicitly trained.
\end{itemize}

\newpage

\end{document}